\documentclass[10pt,twocolumn,letterpaper]{article}

\usepackage[pagenumbers]{cvpr} %

\definecolor{cvprblue}{rgb}{0.21,0.49,0.74}
\usepackage[pagebackref,breaklinks,colorlinks,allcolors=cvprblue]{hyperref}

\def\paperID{21529} %
\def\confName{CVPR}
\def\confYear{2026}

\title{Describe Anything Anywhere At Any Moment}

\author{Nicolas Gorlo \qquad Lukas Schmid \qquad Luca Carlone\\
Massachusetts Institute of Technology\\
{\tt\small \{ngorlo, lschmid, lcarlone\}@mit.edu}
}
\usepackage{amsmath,amssymb,amsfonts}
\usepackage{graphicx}
\usepackage{textcomp}
\usepackage{xcolor}
\usepackage{stfloats}
\usepackage{subcaption}
\usepackage{url}
\usepackage{verbatim}
\usepackage{layouts}
\usepackage{adjustbox}
\usepackage{float}
\usepackage{xspace}
\usepackage{multirow}
\usepackage{pbox}
\usepackage{multicol}
\usepackage{makecell}
\usepackage{array}
\usepackage{colortbl}
\usepackage{booktabs}
\usepackage{tabularx}
\usepackage[absolute,overlay]{textpos}
\usepackage{siunitx}

\NewDocumentCommand{\todo}{o m}{\textcolor{red}{\textbf{TODO\IfNoValueTF{#1}{}{(#1)}:} #2}}

\makeatletter
\def\@gobbletrailingpunct{%
  \@ifnextchar.{\@gobble\@gobbletrailingpunct}{%
    \@ifnextchar,{\@gobble\@gobbletrailingpunct}{%
      \@ifnextchar\space{\@gobble\@gobbletrailingpunct}{}%
    }%
  }%
}
\newcommand{\linkToPdf}[1]{\@gobbletrailingpunct}
\newcommand{\linkToPpt}[1]{\@gobbletrailingpunct}
\newcommand{\linkToCode}[1]{\@gobbletrailingpunct}
\newcommand{\linkToWeb}[1]{\@gobbletrailingpunct}
\newcommand{\linkToVideo}[1]{\@gobbletrailingpunct}
\newcommand{\linkToMedia}[1]{\@gobbletrailingpunct}
\newcommand{\award}[1]{\@gobbletrailingpunct}
\makeatother

\begin{document}

\twocolumn[{%
  \renewcommand\twocolumn[1][]{##1}%
  \maketitle
\begin{center}
    \captionsetup{type=figure}
    \includegraphics[width=.93\textwidth]{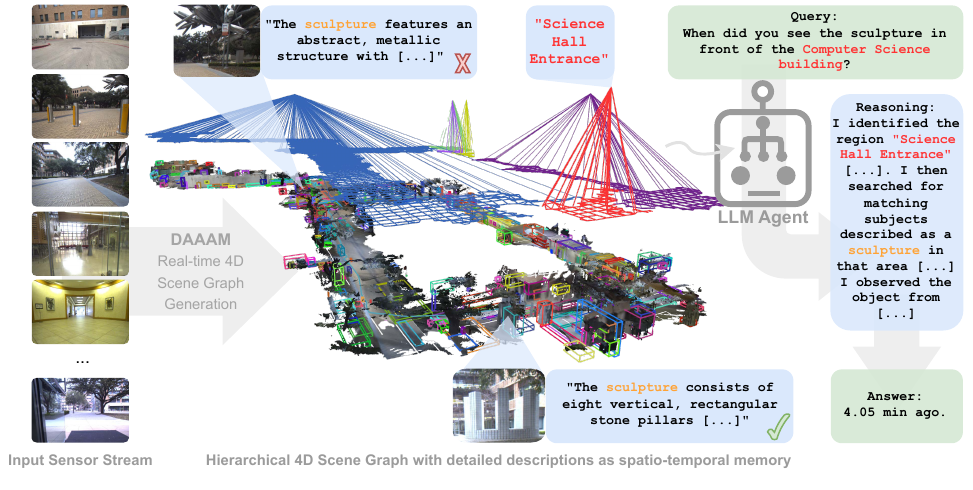}
    \vspace{-1em}
    \captionof{figure}{We present \emph{\textbf{D}escribe \textbf{A}nything, \textbf{A}nywhere, at \textbf{A}ny \textbf{M}oment (DAAAM)}, a real-time, large-scale, spatio-temporal memory for embodied question answering and 4D reasoning. Given RGB-D sensor input \emph{DAAAM} incrementally constructs a hierarchical 4D scene graph with highly detailed annotations that acts as an effective and scalable spatio-temporal memory representation for LLM Agents. %
    }
    \label{fig:title_figure}
\end{center} }]

\begin{abstract}
Computer vision and robotics applications ranging from agumented reality to  robot autonomy in large-scale environments require spatio-temporal memory frameworks that capture both geometric structure for accurate language-grounding as well as semantic detail. 
Existing methods face a tradeoff, where producing rich open-vocabulary descriptions comes at the expense of real-time performance when these descriptions have to be grounded in 3D.
To address these challenges, we propose \emph{Describe Anything, Anywhere, at Any Moment (DAAAM)}, a novel spatio-temporal memory framework for large-scale and real-time 4D scene understanding. %
\emph{DAAAM} introduces a novel optimization-based frontend to infer detailed semantic descriptions from localized captioning models, such as the Describe Anything Model (DAM), 
leveraging batch processing to speed up inference by an order of magnitude for online processing.
It leverages such semantic understanding to build a hierarchical 4D scene graph (SG), which acts as an effective globally spatially and temporally consistent memory representation.
\emph{DAAAM} constructs 4D SGs with detailed, geometrically grounded descriptions while maintaining real-time performance. 
We show that \emph{DAAAM}'s 4D SG interfaces well with a tool-calling agent for inference and reasoning.
We thoroughly evaluate \emph{DAAAM} in the complex task of spatio-temporal question answering (SQA) on the NaVQA benchmark and show its generalization capabilities for sequential task grounding on the SG3D benchmark. We further curate an extended OC-NaVQA benchmark for large-scale and long-time evaluations. %
\emph{DAAAM} achieves state-of-the-art results in both tasks, improving OC-NaVQA question accuracy by 53.6\%, position errors by 21.9\%, temporal errors by 21.6\%, and SG3D task grounding accuracy by 27.8\% over the most competitive baselines, respectively.
We release our data and code open-source.

\vspace{-1em}
\end{abstract}

\section{Introduction}
\label{sec:introduction}

The ability to understand, reason about, and interact with complex, large-scale environments over long time horizons is a crucial challenge in computer vision and a prerequisite to a range of applications in robotics and augmented reality. 
In these applications, perception systems should be able to answer spatio-temporal queries; for instance, a robot operating on a factory floor should be able to answer questions like ``where and when did you last see the red screwdriver?'' or ``can you go and grab the component we assembled last week''. %
However, this requires internal memory representations that i) support spatial reasoning and task planning, ii) extend over long time horizons and large environments, and iii) can be built in real-time with limited computation.

To this end, two main paradigms for spatio-temporal memory systems have emerged.
First, metric-semantic maps ground objects geometrically in 3D reconstructions and semantically lift them~\cite{Jatavallabhula23rss-ConceptFusion,Chen23icra-nlmap,Peng23cvpr-OpenScene3D,Huang23ijrr-avlmaps,Ding23cvpr-pla,Zou25iclr-m3,Zhou24cvpr-HUGS,Shafiullah22arxiv-clipFields,Kerr23iccv-lerf,Yamazaki24icra-openFusion} to act as spatio-temporal memory.
In particular, 3D scene graphs (SGs)~\cite{Armeni19iccv-3DsceneGraphs,Rosinol20rss-dynamicSceneGraphs,Rosinol20icra-Kimera,Wald20cvpr-semanticSceneGraphs,Hughes24ijrr-hydraFoundations} have found widespread interest due to their ability to capture semantic entities and their relationships in a topological graph, grounded in the 3D world.
However, the need for as-expressive-as-possible scene descriptions is fundamentally at odds with the requirement for real-time, mobile computation.
As a result, existing methods either lack semantic detail, using fast but closed vocabulary segmentation or embeddings~\cite{Chen23icra-nlmap,Hughes24ijrr-hydraFoundations,Peng23cvpr-OpenScene3D,Huang23ijrr-avlmaps}, or query large multimodal (MM) models on a per-object basis for highly detailed but very expensive open-vocabulary annotations~\cite{Gu24icra-conceptgraphs,Koch25cvpr-relationfield}.

Alternatively, a second emerging paradigm leverages multimodal large language models (MM-LLMs) to generate detailed scene representations from natural language descriptions. 
Primarily, individual frames or video sequences are annotated by MM-LLMs~\cite{Anwar24icra-remembr,Xie24arxiv-EmbodiedRAG} and stored in a database for later retrieval~\cite{Lewis20neurips-RAG}.
This approach can lead to more expressive representations with notable performance for large scale visual question answering (VQA)~\cite{Anwar24icra-remembr,Xie24arxiv-EmbodiedRAG}.
However, since annotations are stored by frame and not by content, they are often not sufficiently grounded in the 3D world and lack spatial and temporal consistency. For instance, since object observations are not associated and reconciled across frames, these models are typically unable to answer questions involving long-range spatial relationships or object quantities (\textit{e.g.}, ``count the number of chairs'')~\cite{Xie24arxiv-EmbodiedRAG}. 

To address these challenges, we propose \emph{DAAAM: Describe Anything, Anywhere, At Any Moment}.
Our approach, shown in~\cref{fig:title_figure}, builds a hierarchical 4D scene graph (SG) as spatio-temporal memory representation by combining real-time 4D metric-semantic mapping with highly detailed natural-language descriptions for every observed entity.
In particular, we introduce an optimization-based frontend to select key observations and infer detailed descriptions (\textit{e.g.}, using DAM~\cite{Lian25arxiv-DAM}) in batch, speeding up inference time by an order of magnitude for online deployment of large models. 
Our backend then globally optimizes and reconciles the object observations (or \emph{fragments}) created in the frontend, yielding a spatially and temporally consistent 4D SG with description histories for each entity as memory representation.
Finally, a tool-calling agent can efficiently leverage the proposed representation for queries.
We demonstrate \emph{DAAAM} in the complex setting of large-scale spatio-temporal question answering (SQA) on the NaVQA benchmark~\cite{Anwar24icra-remembr} and further show its generalization capabilities in sequential task grounding on the SG3D dataset~\cite{Zhang24arxiv-taskOrientedGrounding}, achieving state-of-the-art results in both cases.
We make the following contributions: 
\begin{itemize}
    \item We introduce \emph{DAAAM: Describe Anything, Anywhere, At Any Moment}, a novel real-time approach to create a 4D SG, an explicit large-scale spatio-temporal memory representation with highly detailed annotations.
    \item We propose an optimization-based method to annotate entities of a 3D scene with large localized captioning models efficiently and online.
    \item We thoroughly evaluate \emph{DAAAM}, achieving state-of-the-art results in spatio-temporal question answering (SQA) and sequential task grounding. We further curate an extended SQA benchmark and release our data and implementation open-source.\footnote{Released upon acceptance.}
\end{itemize}
\section{Related Works}\label{sec:related_works}

Robotic agents need spatial memories that are simultaneously geometrically precise for manipulation, semantically rich for arbitrary natural language instructions, and computationally efficient for real-time operation. 
Existing approaches typically excel in only one of these dimensions.

\textbf{Metric-Semantic Map-based Spatial Memory Systems.} 
Various approaches embed foundation model features~\cite{Radford21icml-clip,Zhai-23iccv-siglip,Oquab23arxiv-dinov2} into different 3D representations: point clouds~\cite{Jatavallabhula23rss-ConceptFusion,Chen23icra-nlmap,Peng23cvpr-OpenScene3D,Huang23ijrr-avlmaps,Ding23cvpr-pla, Kassab24arxiv-bareNecessities}, Gaussian Splats~\cite{Zou25iclr-m3,Zhou24cvpr-HUGS}, radiance fields~\cite{Shafiullah22arxiv-clipFields,Kerr23iccv-lerf} or TSDF volumes~\cite{Yamazaki24icra-openFusion}. 
These approaches often suffer from limited expressiveness, large memory footprint, and inefficient queries (\textit{e.g.}, searching unstructured feature fields in large-scale environments).

Alternatively, 3D scene graphs (SGs)~\cite{Armeni19iccv-3DsceneGraphs,Rosinol20rss-dynamicSceneGraphs,Rosinol21ijrr-Kimera,Wald20cvpr-semanticSceneGraphs}, augment geometric mapping with semantic information and hierarchical structure. 
This was extended to real-time systems~\cite{Hughes24ijrr-hydraFoundations,Wu21cvpr-SceneGraphFusion} for 3D SG construction. 
While these achieve real-time performance, they often rely on closed-vocabulary semantics that limit expressiveness for complex robotic memory queries.
Extensions include temporal changes~\cite{Schmid24rss-khronos, Gorlo24ral-LP2, Looper23icra-3dvsg}, task-grounding~\cite{Maggio24ral-clio,Maggio25arxiv-BayesianFields, Chang25cvpr-ashita}, task-specific visual memory~\cite{Saxena25corl-GraphEQA}, and functional relationships~\cite{Zhang25cvpr-functional3dsg,Koch25cvpr-relationfield}.

An adjacent line of research achieves richer semantic understanding by directly querying MM-LLMs per object~\cite{Gu24icra-conceptgraphs,Koch25cvpr-relationfield}, generating detailed descriptions. 
While semantically rich, these approaches require expensive per-object VLM queries, preventing real-time use. %

Finally, capturing temporal dynamics is important, albeit very challenging for spatial representations. Some systems address this through episodic scene graphs linked across time~\cite{Ginting25corl-mindpalace}, spatio-temporal metric-semantic SLAM handling short- and long-term changes~\cite{Schmid24rss-khronos}, or dynamic point cloud memory with open-vocabulary features~\cite{Liu25icra-dynamem}.

\textbf{View-based Spatial Memory Systems.}
View-based representations directly annotate camera frames with MM-LLMs, sacrificing 3D structure for higher semantic detail.
Retrieval-Augmented-Generation (RAG)~\cite{Lewis20neurips-RAG} approaches store VLM-annotations of frames or short video segments in vector databases~\cite{Anwar24icra-remembr} or hierarchical semantic forests~\cite{Xie24arxiv-EmbodiedRAG}, enabling natural language queries over long horizons. 3D-Mem~\cite{Yang25cvpr-3dmem} retrieves observed frames directly, hoping multimodal LLMs can infer 3D structure from individual views without explicit geometric grounding. Navigation systems ground instructions as textual landmarks in a sequence of per-frame observations~\cite{Shah23corl-lmnav}, use tokenized frame observations~\cite{Zheng24cvpr-navillm} to encode observation history, or use segment-level representations~\cite{Garg24icra-robohop} as a topo-semantic map.
Hybrid approaches aim to combine view-based flexibility with 3D structure~\cite{Hu25arxiv-3dllmMem,Loo25ijrr-openSG} in indoor environments.

These methods offer detailed view-annotations, but their per-frame annotations are not sufficiently grounded in 3D space. Without metric grounding, they struggle with precise spatial reasoning, which is critical, e.g., for manipulation. Lightweight topological approaches~\cite{Garg24icra-robohop,Yokoyama24icra-vlfm} trade semantic richness for efficiency while others require expensive annotations~\cite{Loo25ijrr-openSG}.
In contrast, our approach provides semantic detail alongside geometric grounding and computational efficiency.%

\textbf{VLMs for Spatial Understanding and Task Execution.}
Recent works expand spatial understanding capabilities of VLMs and LLMs~\cite{Ma24arxiv-3dllmsurvey}. Models can reason spatially by training on scene graphs~\cite{Chen24cvpr-spatialVLM,Cheng24arxiv-spatialRGPT}, by using 3D-aware architectures~\cite{Hong23neurips-3dllm,Jia24eccv-sceneverse,Xu25corl-mobilityVLA}, or by training on large scale video data with spatial annotations~\cite{Feng25arxiv-VideoR1}. While these models demonstrate spatial reasoning capabilities, reasoning over large-scale dynamic 3D environments and temporal changes remains challenging.%

For task grounding and execution, systems use scene graphs for hierarchical planning~\cite{Rana23corl-sayplan}, grounding in robotic affordances~\cite{Brohan23corl-saycan}, structured LLM interfaces~\cite{Ray25arxiv-Cypher}, embodied question answering~\cite{Saxena25corl-GraphEQA}, and navigation with chain-of-thought reasoning~\cite{Yin24neurips-sgnav,Yin25cvpr-unigoal}. 
To this end, our contribution provides an efficient, but detailed and general spatial memory that can be used by embodied agents for complex reasoning and task grounding. %
\begin{figure*}
    \centering
    \includegraphics[width=\textwidth]{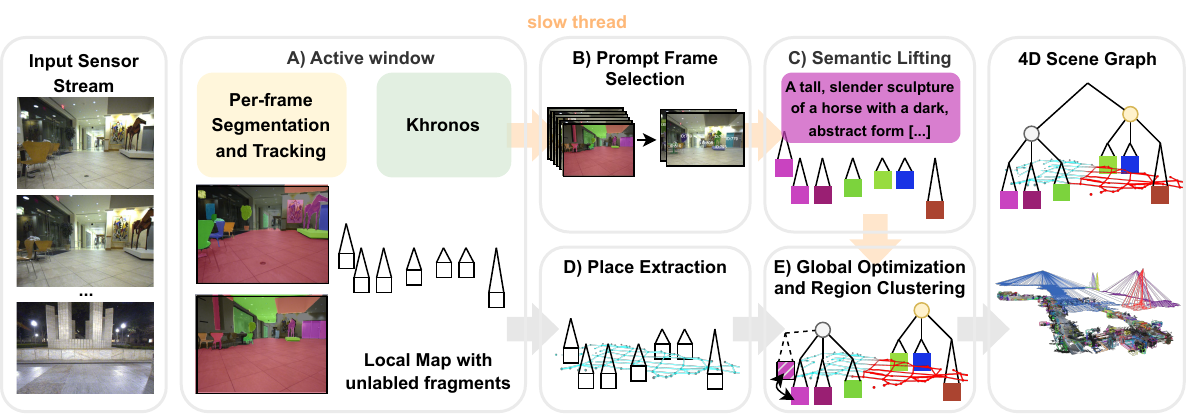}
    \caption{An overview of the proposed approach. Given an RGB-D video stream, we first segment the scene into fragments and track them over time in image space using a lightweight tracker~\cite{Aharon22arxiv-botsort}. %
    We perform metric-semantic mapping using Hydra~\cite{Hughes24ijrr-hydraFoundations} with the Khronos~\cite{Schmid24rss-khronos} frontend on the unlabeled segments to build a 4D map of the environment. %
    To semantically lift the resulting map, we aggregate the tracked observations in parallel and select frames using an optimization-based frame selection algorithm. The selected frames and segments are batch-processed by the Describe Anything Model~(DAM)~\cite{Lian25arxiv-DAM} to generate detailed descriptions for each object. The generated descriptions are finally incorporated back into the map and a 4D scene graph is constructed and clustered into semantically informed regions.
    }
    \label{fig:system_overview}
    \vspace{-10px}
\end{figure*}

\section{Approach}\label{sec:approach}

In the following, we describe \emph{DAAAM: Describe Anything, Anywhere, at Any Moment}. Given RGB-D images and poses as input, our method constructs a 4D scene graph with detailed open-vocabulary semantic annotations in real-time. 
This 4D scene graph acts as a spatio-temporal metric-semantic memory for embodied agents. 
An overview of our pipeline is given in~\cref{fig:system_overview}. 
\emph{DAAAM} is composed of several modules: an active window (A), a parallel thread for semantic lifting of segmentation fragments (B\&C), and modules to construct, maintain and hierarchically structure the large-scale spatio-temporal memory ( D\&E).

\textbf{A) Active Window and Real-time SG Construction.}
We follow the approach of Khronos~\cite{Schmid24rss-khronos} to extract temporally-consistent fragments from the sensor stream in dynamic scenes. 
In particular, we augment Khronos' active window by first splitting each input frame $I_t^{\text{rgb-d}}$ at time $t$ into segments $s_j^t \in \mathbb{R}^{H\times W}$ using Fast-SAM~\cite{Zhao23Arxiv-FastSam} and track them over time using Bot-Sort~\cite{Aharon22arxiv-botsort}.
Each track creates an object fragment $o_j^{0\dots T_j} \in \mathbb{R}^{H\times W\times T_j}$ of $T_j$ observations. 
We then use Khronos to lift each fragment to 3D and reconstruct their shape and position (through time for dynamic objects).
Since geometric segmentation, tracking, and reconstruction is fast, it is run at the sensor rate of 10Hz.

\textbf{B) Prompt Frame Selection.} 
Since extraction of highly-detailed descriptions is an expensive operation, we propose to process only selected frames and annotate fragments in batch.

Specifically, we accumulate fragment observations in consecutive time windows $w_t = [t_{start}, t_{start+m}]$.
After each window, we propose to select frames where each fragment is visible and well-positioned for semantic grounding by the vision language model (VLM).
We formulate the frame selection as a 2-step optimization problem. %
 
Let $\mathcal{O} = \{o_1^w, \ldots, o_m^w\}$ denote the set of tracked object fragments within window $w$, where each $o_j^w$ represents a temporally consistent track across multiple frames. 
For each frame $f_i \in \mathcal{F}^w$ and $o_j^w \in \mathcal{O}$, we define a visibility indicator $v_{ij} \in \{0,1\}$, $v_{ij} := 1$ iff object $o_j^w$ is visible in frame $f_i$, and a view quality score $q_{ij} \in [0,1]$ for object $o_j^w$ in frame $f_i$.

Our selection problem has to balance two competing objectives: minimizing the number of frames sent to the VLM (for computational efficiency) while maximizing the quality of selected fragment-frame pairs (for annotation accuracy). 
To address this, we first solve the set cover problem to find the minimum number of frames $K^\star$ required to observe all object fragments $o_j^w$ at least once:
\begin{equation}
    \begin{aligned}
    K^\star = &\min_{\mathcal{S} \subseteq \mathcal{F}^w} \quad |\mathcal{S}| \\
    \text{s.t.} &\quad \forall o_j^w \in \mathcal{O}: \exists f_i \in \mathcal{S} \text{ with } v_{ij} = 1
    \end{aligned}
    \label{eq:cover}
\end{equation}
We solve \eqref{eq:cover} using a greedy algorithm. Given the minimum frame count $K^\star$, we then solve the binary linear program: %
\begin{equation}
    \begin{aligned}
    \max_{x, y} & \quad \sum_{i=1}^n \sum_{j=1}^m q_{ij} \cdot y_{ij} \\
    \text{s.t.} & \quad \sum_{i=1}^n x_i = K^\star + \epsilon ,
    \quad \sum_{i=1}^n y_{ij} = 1, \\
    &\quad y_{ij} \leq x_i, 
    \text{ } y_{ij} \leq v_{ij}, 
    \text{ } x_i \in \{0,1\}, 
    \text{ } y_{ij} \in \{0,1\} \\
    &\quad \forall i \in [n], 
    \text{ } j \in [m],
    \end{aligned}
    \label{eq:opt}
\end{equation}
where $x_i \in \{0,1\}$ indicates selection of frame $f_i$,  $y_{ij} \in \{0,1\}$ indicates assignment of fragment $o_j^w$ to frame $f_i$, and $\epsilon$ is a slack parameter set to 1. 
The objective maximizes the quality score for each selected fragment, constrained by the conditions that i) the total number of selected frames is $(K^\star+\epsilon)$, ii) if a frame is not selected there is no track in it, and iii) every fragment is assigned to exactly one frame where it is guaranteed to be visible.

While our formulation is applicable to any large model and suitable quality score $q_{ij}$, we design a heuristic $q_{ij}$ that combines position and size components:
\begin{equation}
q_{ij} = \alpha \cdot q_{ij}^{\text{pos}} + (1-\alpha) \cdot q_{ij}^{\text{size}}.
\end{equation}
The position score $q_{ij}^{\text{pos}}$ is the entropy of normalized coordinates to favor centrally located objects, resulting in the maximum value when objects are centered and minimum at frame boundaries. 
The size score $q_{ij}^{\text{size}}$ uses a hyperbolic tangent function that saturates for large objects while penalizing objects below a minimum area threshold $A_{\text{min}}$, ensuring that selected objects are sufficiently visible when being labeled. 
We use $\alpha=0.5$. %

\textbf{C) Semantic Lifting.}
Given the image-fragment pairs from the frame selection algorithm, we batch selected images to annotate all fragments in a single pass of the Describe Anything Model (DAM)~\cite{Lian25arxiv-DAM} to obtain a detailed natural-language description for each fragment. 
We extended DAM with a batch inference strategy that bundles multiple frames and masks into a single tensor, minimizing redundant computation and taking better advantage of parallelization. 
This allows \emph{DAAAM} to run in real-time in real-world environments while still leveraging large, detailed models such as DAM. 
We further assign a CLIP~\cite{Radford21icml-clip} feature and a sentence embedding feature~\cite{Ni21arxiv-sentencet5}, which are similarly obtained in batch, to each fragment to aid semantic search, clustering, summarization, and reconciliation of repeatedly observed objects.
Note that our frame selection algorithm naturally minimizes the number of frames passed to DAM while processing many masks per image, further improving the inference time of the batch-processing. 
In addition, each mask is situated as visible as possible in the image frame leading to high-quality labels.

\textbf{D) Place Extraction.} 
To capture detailed descriptions not only of objects but also of the background, inspired by \cite{Chen22arxiv-LLM2,Honerkamp24ral-MoMa, Maggio24ral-clio}, we further extract \emph{place} nodes $p_j$ in our SG.
Following our approach for objects, we first extract geometric fragments in the active window.
While other methods such as Voronoi diagrams \cite{Hughes24ijrr-hydraFoundations} or sampling \cite{Schmid21ral-volumetricExploration} are also admissible, we extract $p_j$ based on ground \emph{traversability} to capture relevant surfaces and topology.
Traversability is estimated by convolving a robot bounding box with the local volumetric occupancy map maintained in Khronos' active window and squashing along the Z-axis.
We then tesselate this slice into places $p_j$ by inscribing largest traversable rectangles.
To ensure near uniform coverage, each rectangle is subject to a maximum size constraint of 2m.

For semantic lifting, each $p_j$ is first projected to the ground by identifying the surface crossing along the Z-axis in 3D and then projected to all frames that annotate the fragment covering it.
The resulting descriptions and features are assigned by majority voting.
While it may seem intuitive to use full-frame annotations instead of ground-fragment annotations to describe places $p_j$, we find that full-frame queries are often out-of-distribution (OOD) for DAM~\cite{Lian25arxiv-DAM} and therefore use ground annotations.
More detail about places extraction in~\cref{places_appendix}.

\textbf{E) Global Optimization and Region Clustering.}
To achieve a spatially globally consistent memory representation, we continuously optimize the positions of all nodes in our 4D SG in the backend using the factor graph formulation of \cite{Schmid24rss-khronos}.
For temporal consistency, object fragments and place nodes with similar geometry and descriptive features are merged in a reconciliation step. 
To retain temporal information, descriptions of merged objects are appended to form a history, where also the timestamps of the corresponding active window periods are retained.

Finally, we extract regions $R_i$ as hierarchical abstractions by clustering the reconciled low-level SG.
To this end, we first extract regions from the places graph by assigning edge-weights as the cosine distance of the respective semantic features and applying the most-stable-clique finding algorithm of Hydra~\cite{Hughes24ijrr-hydraFoundations}. 
Object nodes are assigned to the closest cluster.
To obtain detailed, representative, and diverse descriptions that summarize the content of regions, we use farthest point sampling of the features of all objects in a region, starting from the mean, and summarize them by prompting a LLM.

\textbf{F) Retrieval-augmented Reasoning} 
For inference, we use a tool-calling agent to answer natural language queries. 
The tool-calling agent has access to tools to a) retrieve objects based on semantic search over the description embeddings, b) retrieve information about the regions and c) retrieve information about the agent. Retrieved information includes spatial and temporal information about each retrieved 4D SG node, making use of the 4D nature of the SG.
More detail about the LLM Agent as well as tool descriptions in~\cref{retrieval_appendix}. %
\section{Experiments}\label{sec:experiments}

We evaluate \emph{DAAAM} on the challenging tasks of spatio-temporal question answering (SQA) as well as sequential task grounding to demonstrate its flexibility and generalization ability. 
We further ablate introduced components and provide run-time analysis for real-time use.
In all evaluations, ground-truth poses are provided as input to all methods. 

\subsection{Spatio-Temporal Question Answering}

\textbf{Dataset.} To assess \emph{DAAAM}'s performance as an explicit large-scale spatio-temporal memory framework, we evaluate it on spatio-temporal question answering (SQA) in the CODa dataset~\cite{Zhang2023arxiv-CODa}, featuring large-scale indoor and outdoor scenes. 
To obtain depth images on the dataset, we use stereo depth estimation~\cite{Wen25cvpr-foundationstereo}.
We adopt the QA-samples from the NaVQA~\cite{Anwar24icra-remembr} benchmark consisting of 210 samples of QA-pairs of different categories, including binary (yes/no) questions, spatial (position) questions , and temporal (time, duration) questions. 
The benchmark further distinguishes different sequence lengths of \emph{Short}, \emph{Medium}, and \emph{Long}, corresponding to average memory durations of \SI{1.2}{\minute}, \SI{4.4}{\minute}, and \SI{12.3}{\minute}, respectively.

\textbf{Baselines.} 
We compare \emph{DAAAM} against recent baselines, including several variations of the state-of-the-art spatio-temporal memory system \textit{ReMEmbR}~\cite{Anwar24icra-remembr}. 
We further compare against a multi-frame VLM, representing a reasoning system without explicit memory, as well as ConceptGraphs~\cite{Gu24icra-conceptgraphs}, a recent open-set metric-semantic mapping-based method.

\textbf{Results.}
We report SQA performance in~\cref{tab:results_remembr_protocol}. 
Our method shows strong performance even when compared to larger-sized models such as ReMEmbR+VILA1.5-13b~\cite{Anwar24icra-remembr,Lin24cvpr-vila} and modern near-real-time methods such as ReMEmbR+NVILA-Lite-2B~\cite{Liu2024arxiv-nvila}, especially for long sequences and temporal reasoning. 
This indicates that geometric structuring of spatio-temporal memory aids scene understanding and the 4D SG representation greatly improves temporal understanding. 
On the other hand, our method struggles with queries which are hard to infer from the limited set of tools (\textit{e.g.}, ``Was the robot driving on the left side of the sidewalk?''), albeit being encoded in the 4D SG, as well as queries that reference small objects.
Note that the binary questions are skewed towards positive labels while LLMs generally predict conservatively, explaining the sub-0.5 question accuracy for some methods (as also observed in~\cite{Anwar24icra-remembr}).

{\definecolor{lightergray}{gray}{0.98}
\newcolumntype{g}{>{\columncolor{lightergray}}c}
\setlength{\tabcolsep}{4pt}
\renewcommand{\arraystretch}{1.7}  %
\setlength{\aboverulesep}{0pt}
\setlength{\belowrulesep}{0pt}
\begin{table*}[]
\caption{Results on the original NaVQA benchmark~\cite{Anwar24icra-remembr}. Results of \textit{ReMEmbR}~\cite{Anwar24icra-remembr} and multi-frame VLM are partially taken from~\cite{Anwar24icra-remembr}, we further ran it with recent language models. Superscript $\dagger$ indicates confounding use of in-context examples in LLM prompts that also appear in the test set. Best (without in-context examples) is bold, best (with in-context examples) underlined if better than without. }
\vspace{-5pt}
\label{tab:results_remembr_protocol}
\adjustbox{width=\textwidth}{%
\begin{tabular}{p{0.2em}lrggggccccgggg}
    \toprule
    \multicolumn{3}{r}{Metric:} & \multicolumn{4}{c}{\cellcolor{lightergray}Descriptive Question Accuracy $\uparrow$} & \multicolumn{4}{c}{Positional Error [m] $\downarrow$} & \multicolumn{4}{c}{\cellcolor{lightergray}Temporal Error [min] $\downarrow$}\\
    \cmidrule(lr){4-7} \cmidrule(lr){8-11} \cmidrule(lr){12-15}
    Method & & Query Length & Short & Medium & Long & All & Short & Medium & Long & All & Short & Medium & Long & All\\
    \midrule
    \multirow{6}{0.2em}{\rotatebox[origin=c]{90}{ReMEmbR~\cite{Anwar24icra-remembr}}} 
    & \multicolumn{2}{l}{NVILA-Lite 2B + GPT-5-mini} & 0.423 & 0.519 & 0.524 & 0.483 & 11.179 & \textbf{23.877} & 65.588 & 39.278 & \textbf{0.313} & 1.240 & 4.472 & 2.518 \\
    & \multicolumn{2}{l}{NVILA-Lite-8B + GPT-5-mini} & 0.551 & 0.605 & 0.714 & 0.607 & 8.486 & 34.409 & 56.297 & 37.337 & 0.363 & 1.491 & 4.979 & 2.840 \\
    & \multicolumn{2}{l}{NVILA-Lite 2B + GPT-5-mini}$\dagger$ & 0.577 & 0.556 & 0.643 & 0.582 & 8.535 & \underline{25.030} & 49.259 & 31.536 & \underline{0.272} & 1.248 & 4.975 & 2.7428 \\
    & \multicolumn{2}{l}{NVILA-Lite-8B + GPT-5-mini}$\dagger$ & 0.538 & \underline{0.630} & 0.571 & 0.582 & 8.467 & 53.279 & 53.717 & 41.485 & 0.436 & 1.336 & 4.122 &  2.414\\
    & \multicolumn{2}{l}{VILA1.5-13b + GPT-4o}$\dagger$ & 0.62 & 0.58 & 0.65 & 0.61 & \underline{5.1} & 27.5 & 46.25 & \underline{29.96} & 0.3 & 1.8 & 3.6 & 2.3 \\
    & \multicolumn{2}{l}{VILA1.5-13b + Llama3.1:8b}$\dagger$ & 0.31 & 0.33 & 0.21 & 0.30 & 159.9 & 151.2 & 165.3 & 159.9 & 9.5 & 7.9 & 18.7 & 13.3 \\
    \arrayrulecolor{lightgray} \cmidrule(lr{.7em}){1-15} \arrayrulecolor{black}
    \multicolumn{3}{l}{Multi Frame VLM | GPT-4o}$\dagger$ & 0.55 & $\times$ & $\times$ & $\times$ & 7.5 & $\times$ & $\times$ & $\times$ & 0.5 & $\times$ & $\times$ & $\times$ \\
    \arrayrulecolor{lightgray} \cmidrule(lr{.7em}){1-15} \arrayrulecolor{black}
    \multicolumn{3}{l}{Concept Graphs~\cite{Gu24icra-conceptgraphs} | NVILA-Lite-8B + GPT-5-mini } & 0.385 & 0.296 & 0.143 & 0.299 & 85.67 & 126.35 & 158.67 & 130.03 & $\times$ & $\times$ & $\times$ & $\times$\\
    \arrayrulecolor{lightgray} \cmidrule(lr{.7em}){1-15} \arrayrulecolor{black}
    \arrayrulecolor{lightgray} \cmidrule(lr{.7em}){1-15} \arrayrulecolor{black}
    \multicolumn{3}{l}{\emph{DAAAM} (Ours) | DAM-3B + GPT-5-mini} & \textbf{0.654} & \textbf{0.630} & \textbf{0.786} & \textbf{0.672} & \textbf{7.282} & 46.015 & \textbf{42.116} & \textbf{33.89} & 0.443 & \textbf{1.030} & \textbf{2.538} & \textbf{1.591} \\    
    \bottomrule
\end{tabular}}
\end{table*}
}

\textbf{Dataset Limitations.}
We find a number of limitations in the NaVQA benchmark that make comparisons tricky.
First, we observed that for several methods in-context examples are provided to the LLM that also appear in the test set, confounding performance.
These methods are marked $\dagger$ in \cref{tab:results_remembr_protocol}, where we additionally evaluate them after removing the confounding examples.
Second, ground truth annotations for spatial questions (e.g., ``Where is the yellow police call pole?'') in NaVQA are annotated as the position from which the object was observed instead of its actual 3D position, strongly favoring view-based memory systems like ReMEmbR. 
Although \emph{DAAAM} is designed to predict actual object positions, we modified it to return the view position for the results in \cref{tab:results_remembr_protocol} to allow a fair comparison on the original NaVQA benchmark.
Third, we find several annotations to be noisy or incorrect (in part leading to the weak performance for Medium spatial queries for \emph{DAAAM}). 
For a dataset of 210 samples, this can greatly skew the results. 
Further, a context window categorizing data into short, medium and long context is provided in which the answer to a question is observed. 
However, we find that for 22 out of 210 samples the provided ground-truth observation time does not fall into the window. 
Finally, the ground truth time and position annotations are calculated from ReMEmbR's video query length, favoring the method, especially for short-horizon questions.

To address these limitations, we re-annotate the spatial queries of the NaVQA dataset with the actual object positions and improve the accuracy of the labels with a custom 3D labeling tool.
We further discard the observation windows and instead evaluate on the entire context of full sequences of the CODa dataset, reflecting more large-scale settings of up to \SI{35.8}{\minute}.
We call this the object-centric NaVQA (OC-NaVQA) dataset and release it publicly.

\textbf{Results on OC-NaVQA.} 
The results on OC-NaVQA are shown in \cref{tab:ocnavqa_results}.
We observe that the 4D SG performance scales well to large-scale long-term settings of up to \SI{35.8}{\minute} and \SI{1.64}{km} of traveled distance, outperforming both metric-semantic map-based spatial memory systems~\cite{Gu24icra-conceptgraphs} and view-based spatial memory systems~\cite{Anwar24icra-remembr}. 
At the scale of the benchmark, ConceptGraphs~\cite{Gu24icra-conceptgraphs} struggles with memory limitations as it maintains a full point-cloud in memory. 
While ReMEmbR~\cite{Anwar24icra-remembr} shows better scaling, we observe limitations in multi-view consistency and spatial reasoning in the large scale setting, leading to reduced performance.

\setlength{\tabcolsep}{4pt} %
\begin{table}[t]
\caption{Results on OC-NaVQA dataset. All models use GPT-5-mini to reason over the constructed memory.}
\vspace{-5pt}
\label{tab:ocnavqa_results}
\adjustbox{width=\columnwidth}{%
\begin{tabular}{lccc}
    \toprule
    Method & \makecell[l]{Question\\ Accuracy $\uparrow$} & \makecell[l]{Positional\\ Error [m] $\downarrow$} & \makecell[l]{Temporal\\ Error [min] $\downarrow$} \\
    \midrule
    ReMEmbR - NVILA-Lite-2B~\cite{Anwar24icra-remembr,Liu2024arxiv-nvila} & 0.432 & 53.466 & 2.287 \\
    ReMEmbR - NVILA-Lite-8B~\cite{Anwar24icra-remembr,Liu2024arxiv-nvila} & 0.463 & 55.894 & 4.106 \\
    \arrayrulecolor{lightgray} \cmidrule(l{.7em}){1-4} \arrayrulecolor{black}
    Concept-Graphs~\cite{Gu24icra-conceptgraphs} & 0.299 & 111.29 & $\times$ \\
    \arrayrulecolor{lightgray} \cmidrule(l{.7em}){1-4} \arrayrulecolor{black}
    \emph{DAAAM} (Ours) & \textbf{0.711} & \textbf{41.75} & \textbf{1.792} \\
    \bottomrule
\end{tabular}}
\end{table}

\subsection{Sequential Task Grounding}

A vital but challenging task in robotics is grounding natural language instructions in the 3D environment. 
We evaluate our method in sequential task grounding on the SG3D~\cite{Zhang24arxiv-taskOrientedGrounding} dataset. 
We follow the evaluation protocol of ASHiTA~\cite{Chang25cvpr-ashita} (Table 2), evaluating on the HM3D~\cite{Ramakrishnan21neurips-hm3d} scenes and using the sequences provided in HOV-SG~\cite{Werby24rss-hovsg} to construct scene graphs.
We compare \emph{DAAAM} against the results reported by Chang~\textit{et al.}~\cite{Chang25cvpr-ashita}, showing sub-task and task accuracy (s-acc and t-acc, respectively) in~\cref{tab:grounding_results}. Task accuracy refers to a full task (composed of multiple subtasks) being correctly grounded in the environment.

We observe significant performance increases over Hydra~\cite{Hughes22rss-hydra}, even when using ground-truth single-word semantic labels \emph{Hydra(GT Seg)}, highlighting the importance of our highly detailed descriptions.
DAAAM further achieves superior results to ASHiTA, a specialized method for hierarchical task analysis, highlighting the generalization ability and flexibility of our 4D SG memory: it represents semantic information in great detail while still providing accurate spatial grounding.

It is worth to point out that SG3D is a semi-synthetic dataset based on HM3D, leading to a real-to-sim gap resulting in a small loss of accuracy for the Describe Anything Model, which is only trained on real data.

\setlength{\tabcolsep}{4pt} %
\begin{table}[]
\caption{Evaluation of Sequential Task Grounding on the SG3D~\cite{Zhang24arxiv-taskOrientedGrounding} benchmark. Our method outperforms metric-semantic map-based spatial meomory systems~\cite{Hughes24ijrr-hydraFoundations,Werby24rss-hovsg} and performs competitively with specialized methods~\cite{Chang25cvpr-ashita}. Results for top 4 methods are taken from Chang et al.~\cite{Chang25cvpr-ashita}.}\label{tab:grounding_results}
\vspace{-5pt}
\centering
\adjustbox{width=.9\columnwidth}{%
\begin{tabular}{lcc}
    \toprule
    Method & s-acc [\%] & t-acc [\%] \\
    \midrule
    Hydra~\cite{Hughes24ijrr-hydraFoundations} + GPT & 8.18 & 2.44 \\
    Hydra~\cite{Hughes24ijrr-hydraFoundations} (GT Seg) + GPT & 14.2 & 6.34 \\
    HOV-SG~\cite{Werby24rss-hovsg} & 8.98 & 1.95  \\
    ASHiTA~\cite{Chang25cvpr-ashita} & 21.7 & 8.78 \\
    \arrayrulecolor{lightgray} \cmidrule(l{.7em}){1-3} \arrayrulecolor{black}
    \emph{DAAAM} (Ours) + GPT & \textbf{22.16} & \textbf{11.22} \\
    \bottomrule
\end{tabular}}
\end{table}

\subsection{Ablation Studies}
\textbf{Language-Augmented Localized Image-Text Retrieval.}
While our explicit descriptions generated using DAM aid in providing context and detail to reasoning LLMs as well as human interpretability, their use for pure retrieval tasks is less clear: One can use a sentence-embedding model to create embedding vectors from the natural language descriptions. This invariably introduces an information bottleneck when comparing to jointly learned embedding spaces produced by VLMs like CLIP~\cite{Radford21icml-clip}.
However, we find some of the retained information to be orthogonal to the information captured by contrastive VLMs and thus concatenate both features in our method.

To test this hypothesis, we conduct a series of retrieval experiments and measure the accuracy of retrieving a localized subject on an image based on the ground-truth description against random subsets of 2500 localized subjects. 
We take five subsets from the Visual Genome dataset~\cite{Krishna16arxiv-visualGenome} and use the entire validation set of the refCOCOg dataset~\cite{Kazemzadeh14emnlp-refCOCO,Yu16eccv-refCOCOg} as a 6th subset (2573 samples). 
Note that for DAM, we run SAM~\cite{Kirillov23iccv-SegmentAnything} to turn the bounding-box annotations into segments, potentially introducing additional error unrelated to the retrieval performance. 
The top-K retrieval accuracy is shown in~\cref{tab:results_retrieval}.

As expected, the retrieval performance of DAM descriptions encoded into a sentence embedding falls short of using a CLIP model for retrieving localized image information. 
However, concatenating the CLIP with the sentence embeddings outperforms CLIP alone, indicating that the two methods capture complementary information. 
Consequently, explicit subject descriptions still aid retrieval tasks, despite the information bottleneck introduced by the text transformation. %

{
\definecolor{lightergray}{gray}{0.98}
\newcolumntype{g}{>{\columncolor{lightergray}}c}
\setlength{\tabcolsep}{4pt}
\renewcommand{\arraystretch}{1.7}  %
\setlength{\aboverulesep}{0pt}
\setlength{\belowrulesep}{0pt}
\begin{table}[]
\caption{Retrieval of localized object descriptions on refCOCOg~\cite{Kazemzadeh14emnlp-refCOCO,Yu16eccv-refCOCOg} and visual Genome~\cite{Krishna16arxiv-visualGenome} datasets.}
\vspace{-5pt}
\adjustbox{width=\columnwidth}{%
\begin{tabular}{p{0.2em}lrggg}
    \toprule
    \multirow{1}{0.2em}{\rotatebox[origin=c]{90}{Data}} & Method & Accuracy [\%] & Top-1 & Top-5 & Top-10 \\
    \midrule
    \multirow{3}{0.2em}{\rotatebox[origin=c]{90}{refCOCOg}} 
    & \multicolumn{2}{l}{CLIP ViT-L/14~\cite{Radford21icml-clip}} & 19.59 & 41.12 & 52.51 \\
    & \multicolumn{2}{l}{DAM-3B~\cite{Lian25arxiv-DAM} w/ Sentence-T5-xl~\cite{Ni21arxiv-sentencet5}} & 18.07 & 39.53 & 51.18 \\
    & \multicolumn{2}{l}{CLIP + DAM w/ Sentence (Ours)}
    & \textbf{25.11} & \textbf{50.10} & \textbf{62.96} \\
    \midrule
    \multirow{3}{0.2em}{\rotatebox[origin=c]{90}{vis. Genome}} 
    & \multicolumn{2}{l}{CLIP ViT-L/14~\cite{Radford21icml-clip}} & 21.64 & 40.04 & 48.4 \\
    & \multicolumn{2}{l}{DAM-3B~\cite{Lian25arxiv-DAM} w/ Sentence-T5-xl~\cite{Ni21arxiv-sentencet5}} & 18.34 & 38.22 & 48.72 \\
    & \multicolumn{2}{l}{CLIP + DAM w/ Sentence (Ours)}
    & \textbf{24.92} & \textbf{47.72} & \textbf{57.76} \\
    \bottomrule
\end{tabular}}
\label{tab:results_retrieval}
\end{table}
}

\textbf{A Case for Explicit Object Descriptions.}
An advantage of our method is the added interpretability, as all nodes of the hierarchical 4D SG are annotated with open-vocabulary natural-language labels. 
This not only aids in human understanding of the spatio-temporal memory, but also helps LLM agents.
In~\cref{tab:ablation_results}, we compare against a baseline ``\texttt{w/o DAM descriptions}'' that, instead of DAM descriptions, only uses visual features and prompts image crops of observed fragments to the MM-LLM agent. 
The results suggest that explicit descriptions aid compositional reasoning of the retrieval-augmented agent, especially for positional and temporal queries. 
When answering binary questions, images appear to provide a better proxy for verification, resulting in superior performance.
We note that our method was designed to run \emph{any} comparatively large model for semantic lifting in parallel. 
Thus, if no explicit descriptions are necessary, a large image embedding model that improves the retrieval accuracy of downstream RAG systems can also be deployed. 

\textbf{Region Clustering.}
To analyze the benefit of our hierarchical region clustering in 4D SGs, \cref{tab:ablation_results} compares against a ``\texttt{w/o region clustering}'' version.
We observe that \emph{DAAAM} benefits from region clustering for all metrics. 
Especially for temporal queries, often focusing on large information context (e.g., ``how much time did you spend inside''), the hierarchical structure of our 4D SG memory appears to facilitate accurate SQA.

\setlength{\tabcolsep}{4pt} %
\begin{table}[]
\caption{Ablation study of \emph{DAAAM} on OC-NaVQA. }
\vspace{-5pt}
\label{tab:ablation_results}
\adjustbox{width=\columnwidth}{%
\begin{tabular}{lccc}
    \toprule
     & \makecell[l]{Question\\ Accuracy $\uparrow$} & \makecell[l]{Positional\\ Error [m] $\downarrow$} & \makecell[l]{Temporal\\ Error [min] $\downarrow$}\\
    \midrule
    \emph{DAAAM} + GPT-5-mini & \underline{0.711} & \textbf{41.75} & \underline{1.792}  \\
    ~w/o~DAM descriptions & \textbf{0.776} & 50.05 & 2.396 \\
    ~w/o~region clustering & 0.707 &  \underline{48.93} & 3.58 \\
    ~w/o~frame selection quality heuristic & 0.627 & 49.92 & \textbf{1.678} \\
    \bottomrule
\end{tabular}}
\end{table}

\textbf{Frame Selection Quality Score.}
``\texttt{w/o frame selection quality heuristic}'' in \cref{tab:ablation_results} highlights the benefit of using the quality heuristic in our optimization based frame-selection algorithm, improving spatial and QA accuracy and performing comparably in terms of temporal accuracy. 
Since temporal queries in the NaVQA benchmark often focus on larger subjects, the quality heuristic may be of lesser importance here.

\subsection{Runtime Analysis}
\label{sec:experiments_runtime}
Since real-time computation is essential for robotics and VR, we evaluate the run-time of DAAAM as an overall system, inference speed-up of our batching strategy, as well as the resulting latency. 
All experiments are performed on CODa~\cite{Zhang2023arxiv-CODa} sequences on a single NVIDIA RTX 5090 GPU.

\textbf{\emph{DAAAM} is a Real-Time System.}
The overall frame rate of our method is compared to several baselines in~\cref{tab:system_timing}.
We observe that, due to \emph{DAAAM}'s efficient architecture and its threaded batch-inference of large models, it can run at the sensor rate of $\SI{10}{Hz}$ in CODa~\cite{Zhang2023arxiv-CODa} even when using comparatively large models such as DAM.
We find the main bottleneck of our method to be input segmentation and tracking, whereas the parallel thread for semantic annotation only becomes a bottleneck for very cluttered or fast-moving scenes with many fragments to annotate. 
Our mean annotation time per fragment is $0.18\pm 0.03$s, thus $5$ new fragments can be annotated per second by a single worker. 
In practice, we find that a single worker is sufficient for the real-time workload of a mobile ground robot in the wild.

In contrast, ConceptGraphs~\cite{Gu24icra-conceptgraphs} is not real-time deployable at the scale of the dataset. ReMEmbR would require down-sampling to a lower frame rate (than the \SI{10}{\hertz} of the CODa dataset), potentially leading to a loss of accuracy.

\begin{table}[t]
\caption{Average frame rate of overall systems for spatio-temporal memory creation [Hz].}\label{tab:system_timing}
\vspace{-5pt}
\adjustbox{width=\columnwidth}{%
\begin{tabular}{cccc}
    \toprule
    & & \multicolumn{2}{c}{RemembeR~\cite{Anwar24icra-remembr}} \\
    \emph{DAAAM} (ours) & Concept-Graphs~\cite{Gu24icra-conceptgraphs} & NVILA-Lite-2B~\cite{Liu2024arxiv-nvila} & NVILA-Lite-8B~\cite{Liu2024arxiv-nvila} \\
    \midrule
    \textbf{11.6} & 0.075 & 4.9 & 4.6 \\
    \bottomrule
\end{tabular}}
\end{table}

\begin{figure}
    \centering
    \includegraphics[width=0.9\columnwidth]{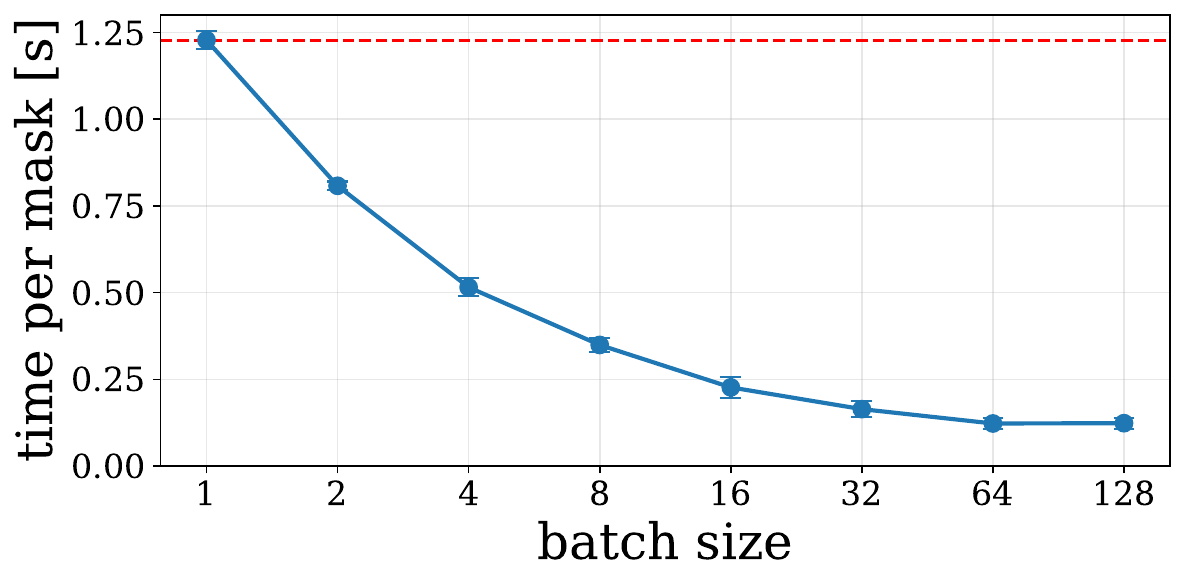}
    \vspace{-5pt}
    \caption{Speedup of DAM~\cite{Lian25arxiv-DAM} inference via batching. Baseline (batch size = 1) dashed red, batch processing solid blue.}
    \vspace{-10pt}
    \label{fig:dam_batching}
\end{figure}

\textbf{Inference Speed Up through Batching.} 
The inference time of DAM for different batch-sizes is shown in \cref{fig:dam_batching}. 
Using our frame selection strategy, we observe notable inference speed-ups.
In our experiments, we use a batch-size between $48$-$128$ (depending on the frame selection outcome), leading to a speed-up by an order of magnitude.

\textbf{Worker Latency.}
We time the frame selection latency at $1.2 \pm 0.74$s and semantic lifting latency at $9.2 \pm 1.4$s, both measuring a full batch. 
The increased latency is an inevitable side-effect of running large models, which delays detailed reasoning about observations by around 10s.
Nonetheless, our experiments demonstrate the benefit of highly detailed memory for robotics and augmented reality, where for most large-scale long-horizon decision making the immediate past may be less important as long as all other observations are accurately summarized.
Thus, throughput appears more important than latency in this case.
Furthermore, the geometric information about all fragments and the background is always maintained in real-time. %
\section{Limitations}\label{sec:limitations}

By the standards of modern large VLMs, the data-body used to train DAM is relatively modest (1.5M samples)~\cite{Lian25arxiv-DAM}. 
Consequently, generated descriptions sometimes fail to capture out-of-distribution objects or uncommon visual features and can hallucinate towards the mean (\textit{e.g.}, predicting elevator doors with handles). 
However, as multimodal LLMs are rapidly evolving, we expect that future detailed localized description models will be able to generate more accurate descriptions and integrate well into DAAAM.

Second, as shown in~\cref{sec:experiments_runtime}, an average of 5.2 new fragments can be annotated per second by a single worker on a Desktop GPU. 
While this suffices for a mobile ground robot, it may be too slow for a dynamic aerial robot or VR headset. 
We note that, since DAAAM aims to run ``comparatively large models'' on a thread with higher latency, smaller models can be run on smaller hardware (thus still comparably large) or at higher throughput.

Finally, although we generate a single spatially and temporally consistent 4D SG by merging nodes, a history of all descriptions is maintained in dynamic nodes which may not scale indefinitely. 
Future work should investigate summarization strategies to keep memory size bounded. %
\section{Conclusion}
\label{sec:conclusion}

We presented \emph{DAAAM}, a novel spatio-temporal memory framework that combines detailed semantic descriptions with geometric grounding for large-scale environments. 
By decoupling geometric tracking from semantic annotation through optimization-based frame selection and batch inference, \emph{DAAAM} overcomes computational constraints of comparatively large vision annotation models.
This enables the real-time construction of hierarchical 4D SGs with highly detailed natural language descriptions. 
Our approach demonstrates substantial improvements over recent methods, achieving state-of-the art results in spatio-tempral question answering and sequential task grounding. 
With real-time performance at 10Hz and scalability to sequences exceeding \SI{35}{min} and \SI{1.5}{km} distance, \emph{DAAAM} provides a practical foundation for embodied agents to understand and interact with complex, large-scale, dynamic environments over an extended time horizon. 
We release our data and code open-source. %
\section*{Acknowledgements}

This work was supported by the the ARL DCIST program and the ONR RAPID program. 
{
    \small
    \bibliographystyle{ieeenat_fullname}

}

\onecolumn
\newpage
\appendix

\renewcommand\thefigure{\thesection.\arabic{figure}}    
\setcounter{figure}{0}

\section{Place Extraction Details}\label{places_appendix}

\begin{figure}[H]
     \centering
     \begin{subfigure}[t]{0.245\textwidth}
         \centering
         \includegraphics[width=\textwidth]{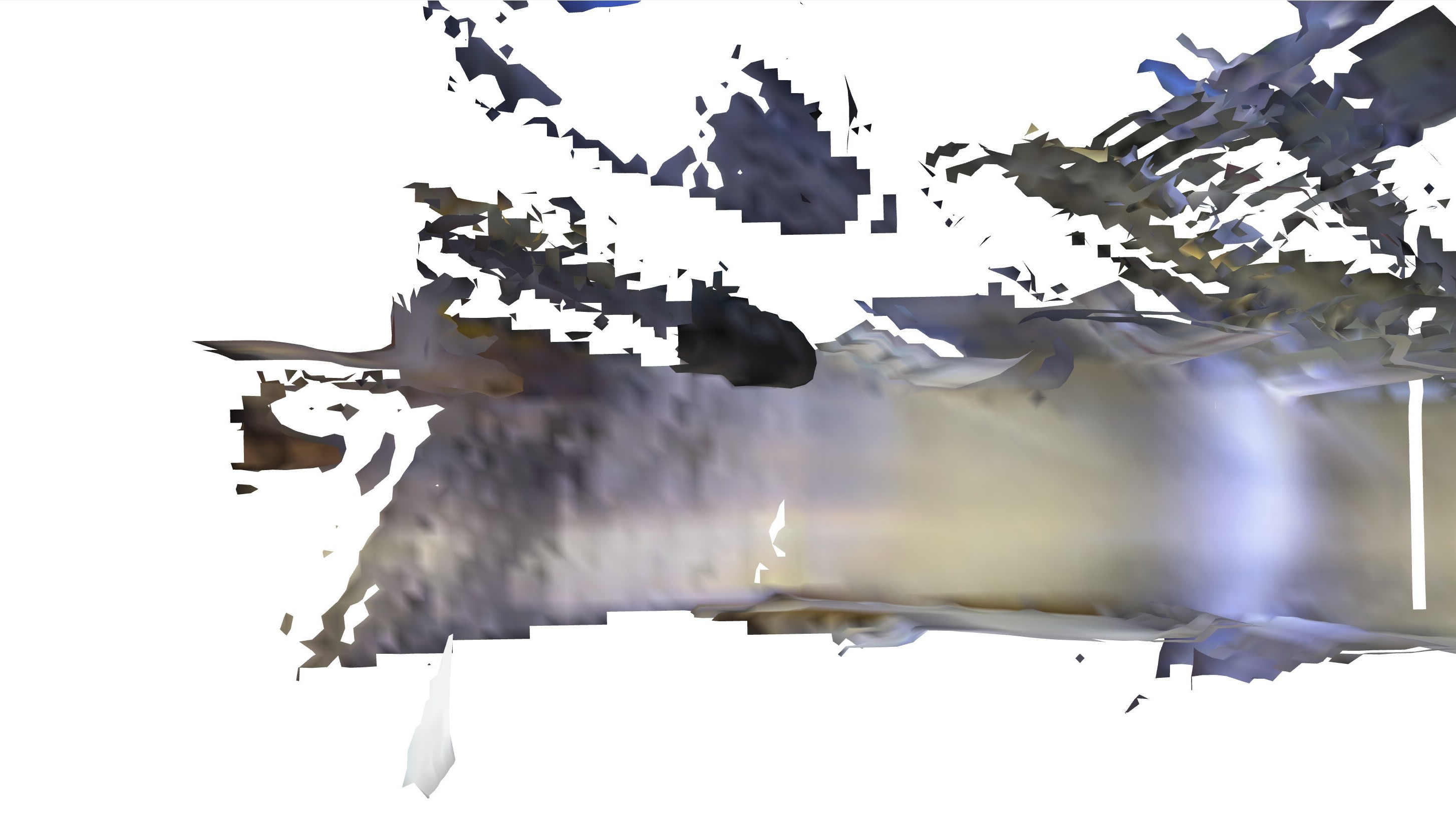}
         \caption{Volumetric 3D reconstruction of the background in the active window of Khronos~\cite{Schmid24rss-khronos}, shown as mesh.}
         \label{fig:places_1}
     \end{subfigure}
     \hfill
     \begin{subfigure}[t]{0.245\textwidth}
         \centering
         \includegraphics[width=\textwidth]{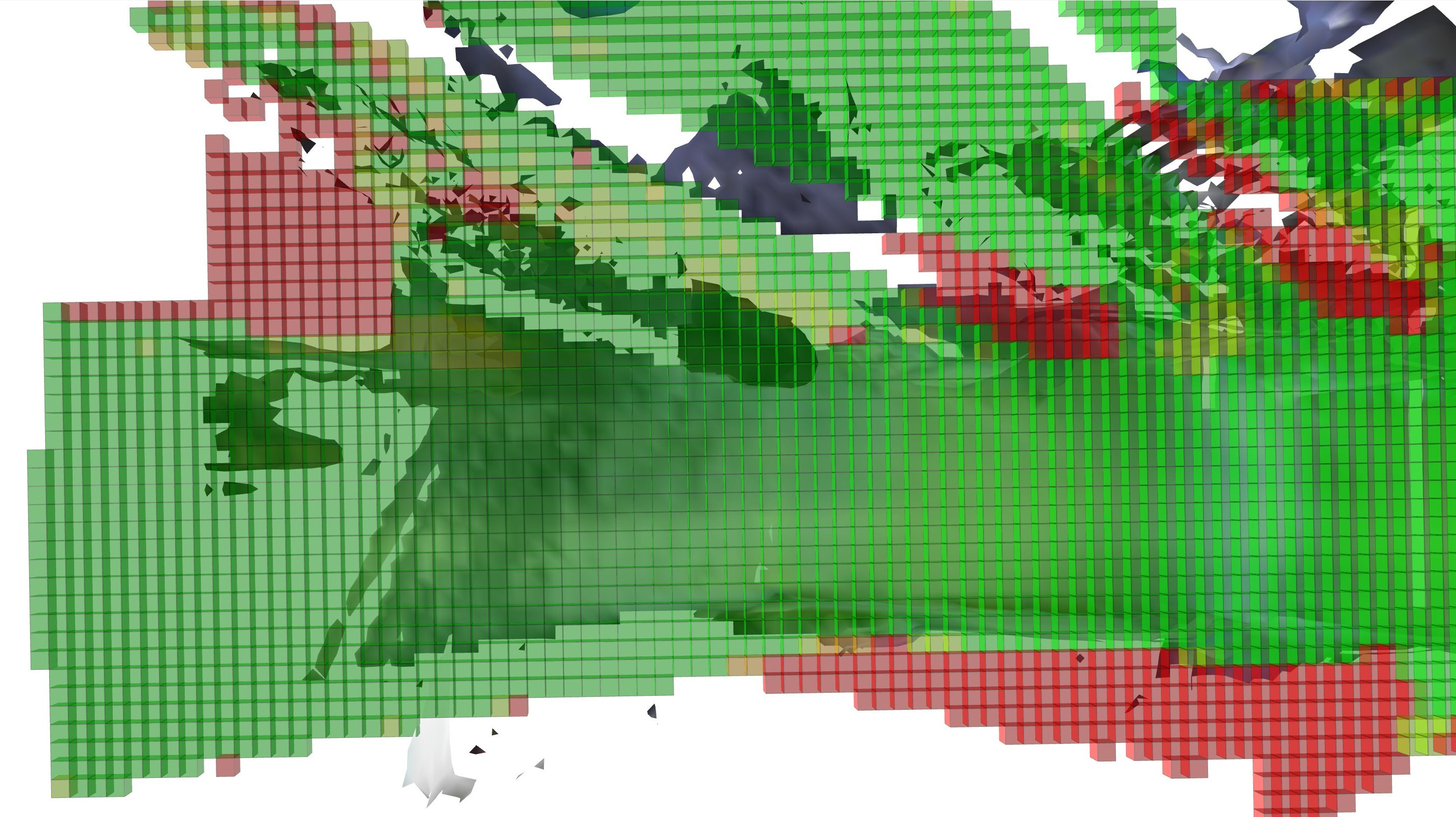}
         \caption{2D traversability field (colored from \textcolor{green}{high} to \textcolor{red}{low} traversability), estimated from the volumetric map in (a).}
         \label{fig:places_2}
     \end{subfigure}
     \hfill
     \begin{subfigure}[t]{0.245\textwidth}
         \centering
         \includegraphics[width=\textwidth]{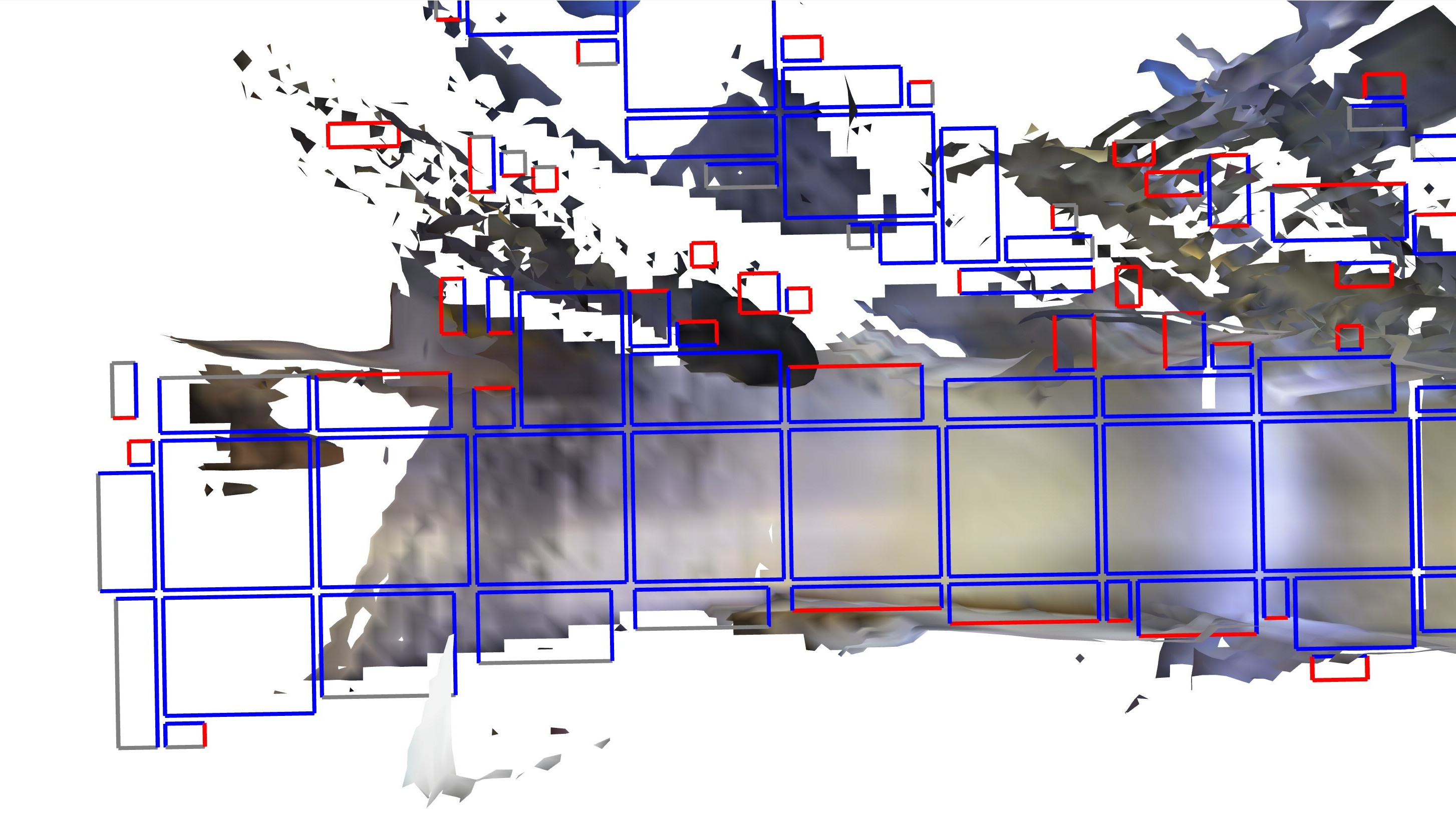}
         \caption{Tesselation of the traversability field into maximal rectangles. Sides shown as bordering \textcolor{blue}{traversable}, \textcolor{gray}{unknown}, and \textcolor{red}{intraversable} areas.}
         \label{fig:places_3}
     \end{subfigure}
    \hfill
     \begin{subfigure}[t]{0.245\textwidth}
         \centering
         \includegraphics[width=\textwidth]{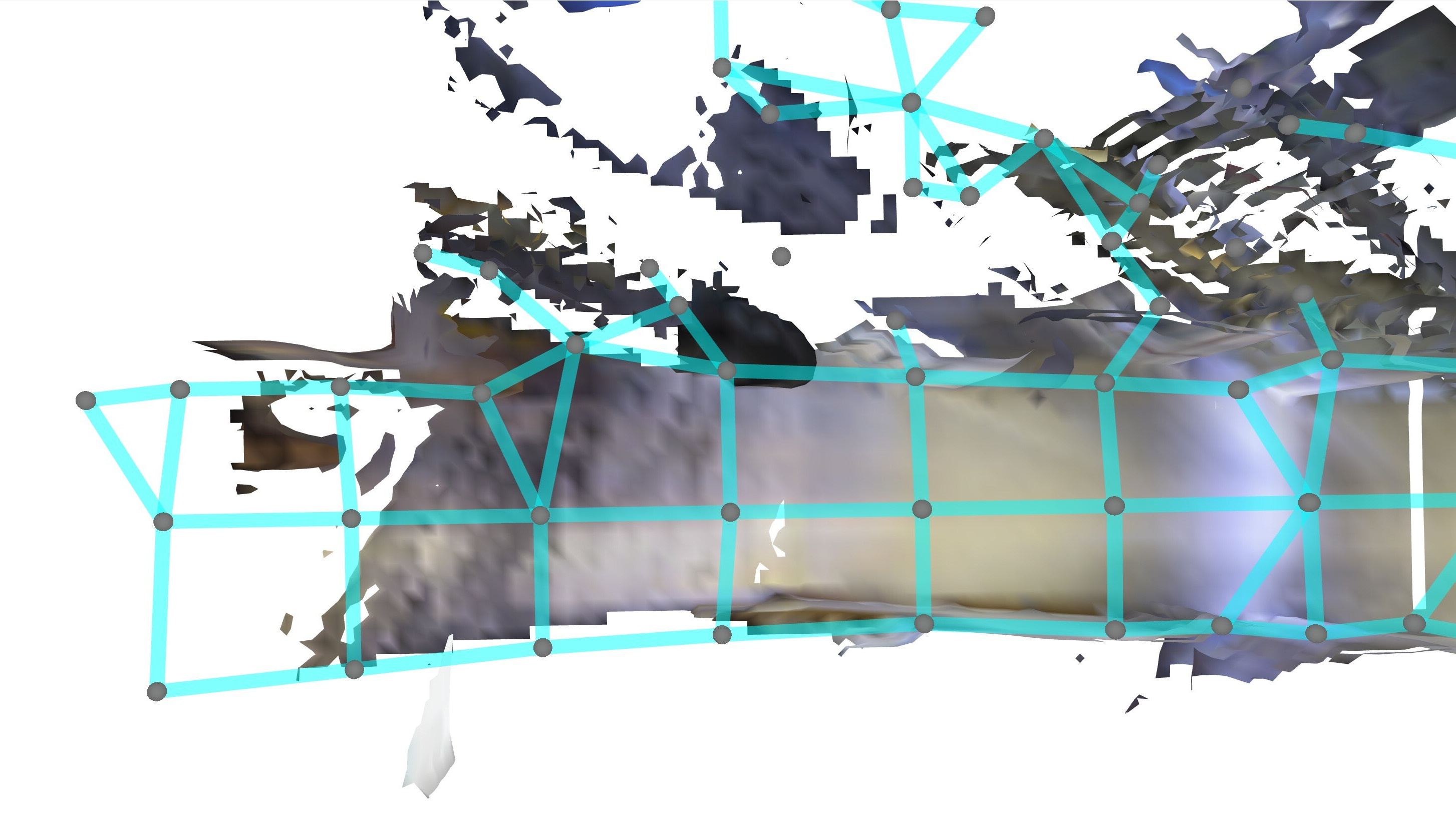}
         \caption{Places graph computed from the centroids and side connectivity information of all rectangles, reflecting the topology of traversable space.}
         \label{fig:places_4}
     \end{subfigure}
        \caption{Illustration of the steps of our traversability place extraction algorithm. The different stages are shown in figures (a)-(d) for a top-down view of the robot moving on a street.}
        \label{fig:places}
\end{figure}

This section further clarifies and illustrates our place extraction approach as introduced in Sec.~\ref{sec:approach}.D).
An overview of the process is shown in \cref{fig:places}.

A volumetric local occupancy map is automatically maintained for 3D reconstruction in the active window of Khronos~\cite{Schmid24rss-khronos}, shown in \cref{fig:places_1}.
The occupancy map is convolved with a bounding box of the robot to compute a 2D traversability field, shown in \cref{fig:places_2}.
The traversability field is then tesselated by inscribing maximal rectangles such that each rectangle contains only traversable space, shown in \cref{fig:places_3}. 
To ensure approximately uniform coverage, we further impose a maximum side length constraint (of $2m$ in this example).
Each side is further classified into bordering traversable, unknown, or nontraversable space.
Finally, the places graph is computed as the centroid of each rectangle in \cref{fig:places_4}.
Two places are considered connected if adjacent sides of the two rectangles are labeled as traversable boundaries.

For semantic lifting, the obtained place nodes are first projected along the Z axis onto the reconstructed floor to obtain a physical surface point, and then projected into frames observing that point to obtain the features as well as semantic descriptions.

\section{Retrieval-augmented Reasoning}\label{retrieval_appendix}
In order for an LLM to interface with the 4D scene graph, we use the LLM as a tool-calling agent. The LLM can use the following tools to retrieve information about the scene graph:
\begin{itemize}
    \item \texttt{semantic\_search}: Given a natural language of a subject, the tool returns the N (=10) most similar fragments based on cosine similarity of the CLIP and sentence embedding features. Returned information includes \texttt{description}, \texttt{position}, and \texttt{observation\_timeline} (list of start- and end-observations).
    \item \texttt{fragments\_in\_radius}: Given a position, the tool returns fragments within a small radius (same information as \texttt{semantic\_search}).
    \item \texttt{region\_information}: The tool returns summaries about the regions in the environment. Information includes the region description, as well as entry and exit positions and times.
    \item \texttt{objects\_in\_region}: Given a region ID and a query description, the tool performs \texttt{semantic\_search} within a region.
    \item \texttt{agent\_trajectory}: Given a start and end position, the tool returns N (=10) equally spaced poses (position + heading) along the agent trajectory.
\end{itemize}

\newpage
\section{Example of frame selection quality-score}
\begin{figure}[H]
    \centering
    \includegraphics[width=0.5\linewidth]{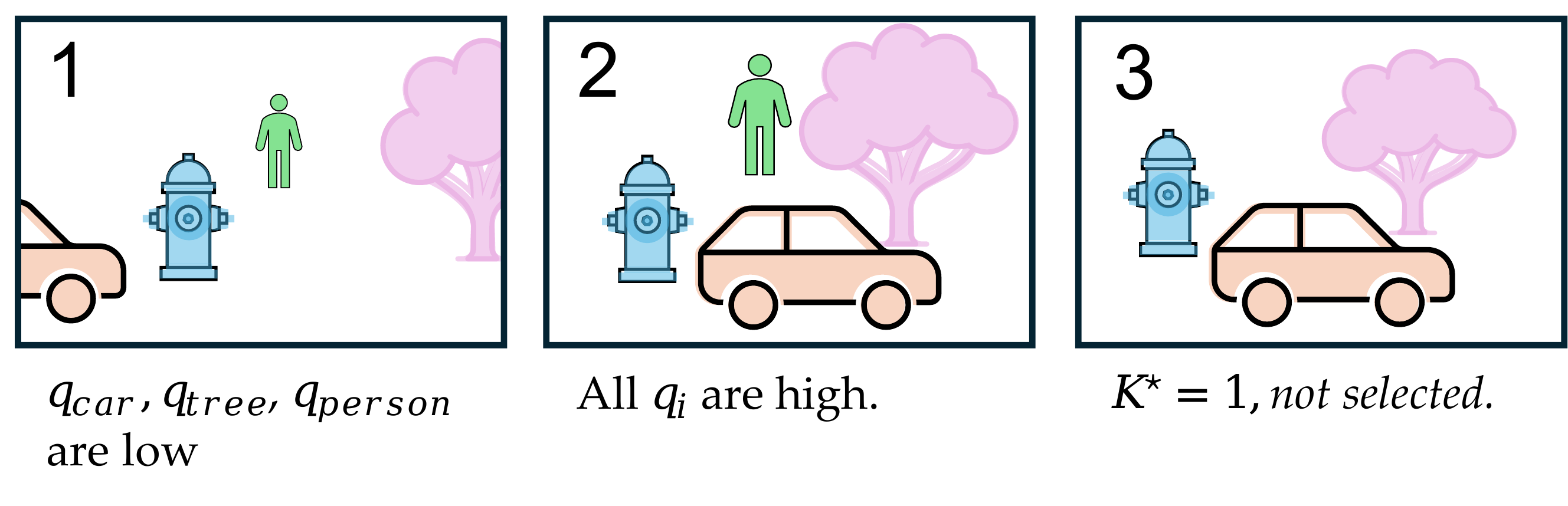}
    \caption{Mock-example for the frame selection heuristic.}
    \label{fig:appendix_frame_selection}
\end{figure}

\cref{fig:appendix_frame_selection} shows a mock-example of the frame selection outlined in~\cref{sec:approach}-B . Given are three sequential images (1-3) with observed fragments \texttt{car}, \texttt{tree}, \texttt{hydrant}, \texttt{person}. Let $\epsilon = 0$.
Then,~\cref{eq:cover} with $K^\star=1$ will eliminate frame $3$ as it does not display all subjects. Thus only a single frame can be selected in ~\cref{eq:opt}.~\cref{eq:opt} will then maximize $q_i$ and therefore select frame $2$ for all fragments \texttt{car}, \texttt{tree}, \texttt{hydrant}, \texttt{person}.

\end{document}